\documentclass[sigconf]{acmart}
\AtBeginDocument{%
  }

\setcopyright{acmlicensed}
\copyrightyear{2025}
\acmYear{2025}
\acmDOI{XXXXXXX.XXXXXXX}
\acmConference[ICMR '25]{ACM International Conference on Multimedia Retrieval}{June 30 -- July 03,
  2025}{Chicago, USA}
\acmISBN{978-1-4503-XXXX-X/2018/06}

\begin{document}

\title{FREAK: Frequency-modulated High-fidelity and Real-time Audio-driven Talking Portrait Synthesis}

\author{Ziqi Ni}
\affiliation{%
  \institution{School of Computer Science and Engineering, Southeast University}
  \city{Nanjing}
  \country{China}
}
\affiliation{%
  \institution{Key Laboratory of New Generation Artificial Intelligence Technology and Its
Interdisciplinary Applications, Ministry of Education, China}
\country{}
}
\email{zqni@seu.edu.cn}

\author{Ao Fu}
\affiliation{%
  \institution{School of Computer Science and Engineering, Southeast University}
  \city{Nanjing}
  \country{China}
}
\affiliation{%
  \institution{Key Laboratory of New Generation Artificial Intelligence Technology and Its
Interdisciplinary Applications, Ministry of Education, China}
\country{}
}
\email{220232248@seu.edu.cn}

\author{Yi Zhou}
\authornote{Corresponding author}
\affiliation{%
  \institution{School of Computer Science and Engineering, Southeast University}
  \city{Nanjing}
  \country{China}
}
\affiliation{%
  \institution{Key Laboratory of New Generation Artificial Intelligence Technology and Its
Interdisciplinary Applications, Ministry of Education, China}
\country{}
}
\email{yizhou.szcn@gmail.com}



\begin{abstract}
Achieving high-fidelity and accurate lip-speech synchronization in audio-driven talking portrait synthesis is particularly challenging. Some studies utilize multi-stage pipelines or diffusion models for high-quality talking portraits; however, they suffer from excessive computational costs. Some approaches achieve remarkable progress on specific individuals with low resource requirements, yet still exhibit mismatched lip movements. The aforementioned methods are modeled in the pixel domain. We observed that there are noticeable discrepancies in the frequency domain between the synthesized talking videos and natural videos. Currently, no research on talking portrait synthesis has considered this aspect.
To address this, we propose a \textbf{FRE}quency-modulated, high-fidelity, and real-time \textbf{A}udio-driven tal\textbf{K}ing portrait synthesis framework, named \textbf{FREAK}, which models talking portraits from the frequency domain perspective, enhancing the fidelity and naturalness of the synthesized portraits. FREAK introduces two novel frequency-based modules: 1) the Visual Encoding Frequency Modulator (VEFM) to couple multi-scale visual features in the frequency domain, better preserving visual frequency information and reducing the gap in the frequency spectrum between synthesized and natural frames. and 2) the Audio Visual Frequency Modulator (AVFM) to help the model learn the talking pattern in the frequency domain and improve audio-visual synchronization. Additionally, we optimize the model in both pixel domain and frequency domain jointly.
Furthermore, FREAK supports seamless switching between one-shot and video dubbing settings, offering enhanced flexibility. Due to its superior performance, it can simultaneously support high-resolution video results and real-time inference. 
Extensive experiments demonstrate that our method synthesizes high-fidelity talking portraits with detailed facial textures and precise lip synchronization in real-time, outperforming state-of-the-art methods. 

\end{abstract}

\ccsdesc[500]{Computing methodologies~Biometrics}
\ccsdesc[300]{Computing methodologies~Image representations}
\ccsdesc[300]{Computing methodologies~Reconstruction}

\keywords{Talking Portrait Synthesis, Video Dubbing, Frequency Domain Analysis, High-Quality Face}


\maketitle

\begin{figure}
    \centering
    \includegraphics[width=1\linewidth]{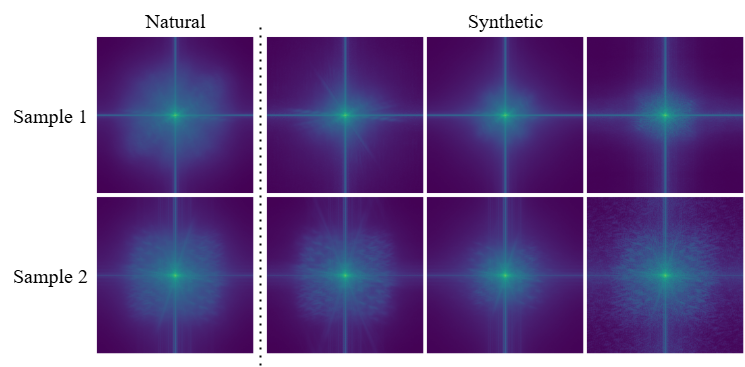}
    \caption{Frequency analysis on natural and synthetic videos. The leftmost column shows the averaged FFT spectrum of two distinct natural videos, while the right three columns (from left to right) display the averaged FFT spectra of corresponding synthetic videos generated by RAD-NeRF \cite{radnerf}, DINet \cite{dinet}, and SadTalker \cite{sadtalker}, respectively. The two rows represent data from talking videos of two distinct identities.}
    \label{fig:freq-compa}
\end{figure}
\section{Introduction}
Audio-driven talking portrait synthesis has gained significant attention in recent years \cite{emo,hallo,radnerf,adnerf,stablesync,segtalker,difftalk}, due to its broad range of applications in areas such as digital avatars, online meetings, live streaming and movie production. The key challenges of this task are: 1) synchronizing audio and lip movements, 2) maintaining realistic and high visual quality, and 3) ensuring real-time performance across various application scenarios.

There are many attempts to realize high-fidelity talking portrait.
Recent end-to-end methods directly map speech spectrograms to video frames leveraging different intermediate representations \cite{everybodytalk,styletalk,aniportrait,makeittalk,iplap,lsp,sadtalker}. Lu et al. \cite{lsp} leverage facial landmarks as intermediate motion representation and an image-to-image translation network to realize real-time photorealistic talking head animation.
Some talking head generation methods \cite{wav2lip,musetalk,dinet,stablesync} without intermediate representations adhere to a mask-modeling paradigm, where the model takes reference frames, masked frames, and audio as inputs to generate frames with synchronized lip movements. 
More recently, some methods \cite{hallo,emo,aniportrait,difftalk,diffusedhead} employ diffusion models and multi-stage pipelines to ensure output quality, these approaches are highly time-consuming.
Besides, a subset of approaches \cite{lsp,radnerf,ernerf,daetalker} focuses on personalized training to achieve high-fidelity and high-quality talking head videos. RAD-NeRF \cite{radnerf} models a 3D talking head with NeRF \cite{nerf}, which not only enhances video quality through geometric regularization but also significantly reduces training resource consumption and improves runtime efficiency. Du et al. \cite{daetalker} employs a diffusion autoencoder to learn implicit personalized latent representations, which are subsequently decoded into talking videos through a DDIM decoder.

However, we observe that the synthetic videos obtained by these methods still exhibit visually perceptible discrepancies in the frequency domain compared to natural videos, as illustrated in Fig. \ref{fig:freq-compa}. In recent studies on the diagnosis of deep neural networks, frequency domain analysis has been proved to be an effective tool \cite{spectralbias}. Some works \cite{freqawaredetection,freqtwobranch} have tried to utilize the frequency discrepancies between generated images and natural images for deepfake detection. Nevertheless, most talking portrait synthesis methods mainly focus on the lip-sync and talking face in the pixel domain, thereby limiting the realism and fidelity of their results.

In this paper, we introduce FREAK, a mask-modeling-based framework with frequency-modulation for high-fidelity and real-time audio-driven portrait synthesis. Two novel frequency-based modules are involved. The Visual Encoding Frequency Modulator (VEFM) adaptively couples multi-scale visual features from two different types of visual information (reference frame and masked frame). By operating in the frequency domain, this approach helps the model learn to globally select key frequency patterns from the reference frame and combine them with the spectral features of the masked frame. This method preserves frequencies better compared to operations performed after concatenation in the pixel domain. Additionally, we integrate visual information into the audio embeddings to strengthen the correlation between audio and visual features. To further align the visual patterns with the audio embeddings, we introduce a Audio Visual Frequency Modulator (AVFM).
This module selectively modulates the visual spectral features with audio embeddings, ensuring that the generated videos are closely aligned with the input audio. Together, these components and frequency regularization work synergistically to improve the naturalness and fidelity of the generated talking videos.

Moreover, we observe that many existing methods \cite{sadtalker,hallo,emo,aniportrait,diffusedhead} either only support one-shot setting, preventing frame-by-frame lip replacement for applications such as video dubbing and translation, or they \cite{wav2lip,emogen,difftalk,segtalker,musetalk} only perform lip replacement without capturing overall head motion when provided with a single reference image. We aim to develop a framework that can flexibly switch between one-shot and video dubbing settings, significantly enhancing its practicality and versatility. The implementation idea is relatively simple: following the mask-modeling paradigm, the model accept two types of mask inputs. In the one-shot setting, the entire head is masked, allowing the model to synthesize a head appearance synchronized with the audio. In the video dubbing setting, the lower face is masked, enabling the model to synthesize only the matching lip movements.

In the meantime, most methods operate at a relatively low resolution (e.g., $256\times256$). Scaling them up to higher resolution videos either significantly increases computational costs or leads to suboptimal results.
The efficiency of the entire framework enables real-time generation of high-resolution and high-fidelity videos. To the best of our knowledge, FREAK is the first to introduce frequency-domain learning in talking head generation.

The contributions of this work are summarized as follows:
\begin{itemize}
    \item We introduce two novel frequency-domain-based modules. A visual feature modulator adaptively couples multi-scale reference image features and masked image features in the frequency-domain to enhance the output's quality and fidelity. A cross-modal fusion module modulates the visual feature in the frequency domain using the audio feature, where the audio serves as a frequency filter to better align lip movements.
    \item We introduce a novel talking portrait synthesis framework FREAK that seamlessly integrates video dubbing and one-shot generation.
    \item Extensive experiments demonstrate the effectiveness of FREAK, highlighting its ability to achieve high-fidelity and high-resolution talking portrait generation in real-time.
\end{itemize}

\section{Related Work}
In this section, we briefly review the work related to the proposed
model. We mainly focus on the following topics: audio-driven talking portrait generation and frequency domain learning.
\subsection{Audio-driven Talking Portrait Synthesis} 
Audio-driven talking portrait synthesis encompasses a variety of methods. 
A significant portion of methods \cite{avcat,wav2lip,stablesync,difftalk,diff2lip,musetalk,iplap,segtalker,talklip} adhere to the mask-modeling paradigm. These methods mask the original video frames, typically masking the lower face or mouth, and randomly select reference frames that provide identity information to input into the model, thereby generating video frames synchronized with the audio.
Some of these methods \cite{makeittalk,sadtalker,aniportrait,iplap,lsp} employ intermediate representations, such as facial landmarks or 3D Morphable Models (3DMM) \cite{3dmm}. They usually first learn the mapping from audio to the intermediate representation and then render the final video frames by combining the intermediate representation and reference frame. Introducing intermediate representations can reduce the difficulty of network learning and ensure the quality of the generated videos. It also enables one-shot generation using the intermediate representation. For instance, Zhong et al. \cite{iplap} employ a Transformer Encoder to first learn the mapping from audio to the lower-half facial landmarks. And then combine the masked video frames with reference frame images and their corresponding landmarks through a translation module to obtain the final video frames. However, incorporating intermediate representations may accumulate errors and lead to relatively stiff facial expressions.
More recent methods \cite{loopy,emo,hallo} utilize powerful diffusion models to achieve superior visual effects and enhanced generalization capabilities. Jiang et al. \cite{loopy} directly input a single reference image and audio into a diffusion model, which can iteratively generate video frames without even requiring intermediate representations or mask modeling. However, such methods have extremely high demands for both training data scale, computational resources and inference time. In addition, there is a class of methods \cite{radnerf,ernerf,lsp,daetalker,damc} with lower computational resource requirements for specialized person training. These methods typically employ 3D technology, although some incorporate techniques from 2D-based methods. For example, Lu et al. \cite{lsp} utilize intermediate representations and renders results through a translation network, while Du et al. \cite{daetalker} employ a Diffusion Auto-encoder to generate talking portrait videos. Compared to other categories of methods, they offer higher fidelity and inference speeds.

We aim to combine the advantages of the above methods and develop a talking portrait synthesis framework with high-fidelity, high-resolution, and real-time performance from the frequency domain perspective, which has been ignored by previous works.

\subsection{Frequency Domain Learning} 


Frequency domain analysis has long been a fundamental and effective tool in traditional signal processing \cite{dip}. In deep learning, frequency-domain methods have also been widely employed for various tasks. The frequency domain provides a non-local \cite{globalfilter,adaptivefourier} or domain-generalizable \cite{freqdg} representation, rich in information, and offers a distinct perspective compared to spatial domain representations. 
Tatsunami et al. \cite{dynamicfilter} introduced a Fast Fourier Transform (FFT)-based token mixer to efficiently replace multi-head self-attention. Similarly, Huang et al. \cite{aff} achieved computationally efficient global learning by incorporating adaptive frequency filters, as opposed to large convolution kernels, self-attention mechanisms, or fully connected layers.
Wang et al. \cite{freqmotionmag} decoupled motion field in the frequency domain to achieve video motion magnification, while Gao et al. \cite{freqderain} combined frequency domain learning with contrastive constraints to achieve image de-raining.
Additionally, many previous studies \cite{focalfreqloss,ssdgan,freqmotion} about image and video generation have explored the effectiveness of the frequency-domain analysis. Yang et al. \cite{freqmotion} explore two novel frequency-based regularization modules to improve the appearance of the person and temporal consistency between adjacent frames in human motion transfer task. Chen et al. \cite{ssdgan} propose a frequency-aware classifier to address the issue of high frequencies missing in the discriminator of standard GAN.
In the domain of deepfake detection, there is also a class of frequency-based methods\cite{freqdeepfake,freqtwobranch,freqawaredetection,realtimedeepfake}. Lanzino et al. \cite{realtimedeepfake} introduce a real-time deepfake detection method using Binary Neural Networks (BNNs) combined with Fast Fourier Transform (FFT) and Local Binary Pattern (LBP) as multi-domain features to identify image manipulations in both frequency and texture spaces. Tang et al. \cite{freqawaredetection} introduce a novel frequency-aware approach called FreqNet, centered around frequency domain learning, specifically designed to enhance the generalizability of deepfake detectors. However, as far as we know, in the research field of talking head generation, no prior work has explored the integration of frequency domain learning. Our work will fill this gap and introduce a new direction with significant potential to drive future research about talking portrait generation. 

\begin{figure*}[!t]
    \centering
    \includegraphics[width=1\linewidth]{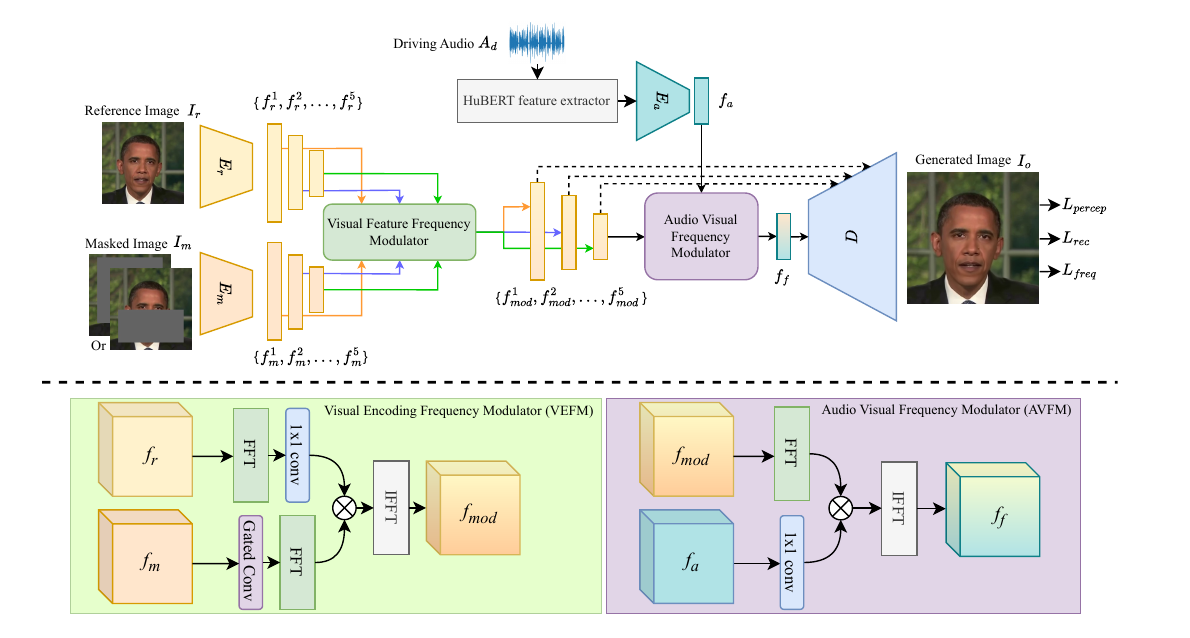}
    \caption{The framework of our method. The model takes a single reference image, an audio clip, and a masked image (which can be either a half mask or a full mask) as inputs, and produces a frame that is synchronized with the driving audio. The lower half of the figure presents the detailed structure of the two frequency-domain modulation modules.}
    \label{fig:framework}
\end{figure*}

\section{Method}
In this section, we mainly introduce the proposed audio-driven talking portrait synthesis framework (FREAK) in detail. We first describe the overall framework of FREAK, and then, we introduce the details of the proposed modules and loss functions for optimization.

\subsection{Overview}
The framework is depicted in Fig. \ref{fig:framework}. The entire model consists of two visual encoders for the reference image and masked image respectively, an audio encoder, and a decoder. Inspired by \cite{mobilenet}, both encoders and the decoder are constructed with Inverted Residual Blocks. For the features from visual encoders are modulated via VEFM. Subsequently, the modulated visual features and audio features are further modulated by AVFM before being fed into the decoder. The modulated features are connected to the decoder in a U-Net-like manner.

We first use a pre-trained Hubert feature extractor \cite{hubert} to extract features from the audio $A_d$, obtaining the Hubert features $f_{hu}$. In the one-shot setting, $f_{hu}$ corresponding to different frames are sequentially fed into the network, along with the reference image $I_r$, and the image $I_m$ with the entire head masked. This setup generates a talking face portrait with natural head movements using a single image. In the video dubbing, the reference image is replaced with each frame of the original video, and the corresponding half-masked image is used as the masked input. Specifically, the extracted Hubert features are passed through an audio encoder to obtain the audio feature $f_a$. Separate visual encoders, $E_r$ and $E_m$, extract multi-scale visual feature maps, $f_r^{i}$ and $f_m^{i}$, respectively, from both reference image and masked image, where $i \in {1, 2, ..., t}$, and $f^i$ represents the $i$-th feature map from the encoder. The visual feature maps are fused in the frequency domain. The final modulated visual feature $f_{mod}^t$ is then modulated through the audio feature $f_a$, and fed into the decoder, while the remaining visual features are concatenated to the decoder via skip connections. Finally, the decoder generates the video frame $I_o$ with synchronized lip movements.

\subsection{Frequency Domain Modulation}
\textbf{Visual Encoding Frequency Modulator.} 
The core idea of Visual Encoding Frequency Modulator (VEFM) is to adaptively couple the masked features, which lack lip or face information, with the reference feature spectrum containing global information.
In the frequency domain, each point corresponds to a global pattern in the spatial domain, thus the network focuses on learning the global structure and avoids copying spatial local features.

Furthermore, directly manipulating visual features in the spatial domain makes it difficult for the network to learn high-frequency patterns, resulting in a sharp degradation of high-frequency components in the output \cite{spectralbias}. This is also why previously synthesized videos often suffered from frequency insufficiency. By coupling the two types of visual features in the frequency domain, it is possible to well preserve the frequencies from natural frames and enable the network to adaptively learn how to combine and manipulate frequency patterns.


Specifically, as illustrated in Fig. \ref{fig:framework}, The majority of information in the visual part comes from the reference image, so it is necessary to retain as much of the spectrum from the reference image as possible. Therefore, we first perform a 2D Fast Fourier Transform (2D-FFT) to convert it the reference features $f_r$ into the Fourier space. The real part is then passed through a $1\times 1$ convolution to learn adaptive channel-wise weighting factors, forming the filter $H_r$:
\begin{equation}
H_r = \text{Conv}(\text{Re}(\mathcal{F}(f_r)))
\end{equation}
where $ \mathcal{F} $ represents the 2D-FFT, $ \text{Re}(\cdot) $ extracts the real part of the frequency spectrum, and $\text{Conv}$ is the $1 \times 1$ convolution operation. Here, the $1\times1$ convolution serves to adjust the frequencies of different channels and ensure that the number of channels matches that of the masked frequency features, facilitating subsequent multiplication operations.

For the masked features, since the features from the masked image have most of their spatial information missing, we first apply a gated convolution ($\text{GC}$) \cite{gatedconv} to selectively extract spatially meaningful features, which are then transformed into the frequency domain, denoted as $F_m$:
\begin{equation}
F_m = \mathcal{F}(\text{GC}(f_m))
\end{equation}

Modulation is performed through element-wise multiplication between the masked feature spectrum and the filter derived from the reference features, producing the modulated visual features. The modulated features are then transformed back into the spatial domain through the inverse fast Fourier transform, obtaining $f_{mod}$:
\begin{equation}
f_{mod} = \mathcal{F}^{-1}(F_m \odot H_r)
\end{equation}
where $\odot$ denotes element-wise multiplication and $\mathcal{F}^{-1}$ represents the inverse fast Fourier transform.

The following experiments demonstrate the effectiveness of VEFM.



\begin{table*}[htbp]
\caption{Quantitative Evaluation With The State-of-art Methods}
\begin{center}
\begin{tabular}{|c|c|c|c|c|c|c|c|c|c|}
\hline
\textbf{Method/}&\textbf{Video} &\multicolumn{5}{|c|}{\textbf{Reconstruction}}&\multicolumn{2}{|c|}{\textbf{Cross Audio}} &\textbf{Runtime}\\
\cline{3-9} 
\textbf{Score} &\textbf{Resolution} &\textbf{FID↓} &\textbf{FVD↓}& \textbf{LSE-D↓}& \textbf{LSE-C↑} & \textbf{LMD↓} &\textbf{LSE-D↓} &\textbf{LSE-C↑}&\textbf{(ms)}\\
\hline
Ground Truth &$512\times512$& 0 & 0 & 6.636 & 8.689 & 0 & 6.759&7.683
 & / \\
Wav2Lip\cite{wav2lip}& $96\times96$ & 32.714& 4.945 & \textbf{5.973} & \textbf{9.544} & 2.536 & \textbf{6.619}	
& \textbf{9.095} & \textbf{8.146}\\
DINet\cite{dinet} & $256\times256$ & 63.600 &4.497& 7.154 & 8.025 & 2.762 &8.294&6.828
& 166.384\\
IP-LAP\cite{iplap} & $256\times256$ & 19.500 &3.923& 8.077 & 6.663 & 2.782 & 9.901&4.752
&190.356 \\
MuseTalk\cite{musetalk} & $256\times256$ & 12.374 &2.168& 8.186 & 6.696 & 2.548 & 10.933&3.879
&68.595\\
RAD-NeRF\cite{radnerf} & $512\times512$ &69.761&4.920 &9.254&5.733 &2.432 & 9.870 & 5.052 & 52.532\\
Ours & $512\times512$ & \textbf{10.514}& \textbf{1.896} & 7.202 & 7.972 & \textbf{1.800} & 8.684 & 6.387
&12.727\\
\hline
SadTalker(1-shot)\cite{sadtalker}&$256\times256$ &108.961 &4.271 & 7.984 & 7.129 & 4.106 & 7.931&7.459
& 50.908 \\
Ours(1-shot) &$512\times512$ &\textbf{61.860} &\textbf{3.728}& \textbf{7.410} & \textbf{7.569} & \textbf{3.258} & 8.509&6.566
 & \textbf{12.913} \\
\hline
\end{tabular}
\label{quan}
\end{center}
\end{table*}

\textbf{Audio Visual Frequency Modulator}. 
The Audio Visual Frequency Modulator (AVFM) operates similarly to VEFM, but it utilizes audio features to modulate visual features in the frequency domain, with the aim of improving lip synchronization. Many prior methods \cite{wav2lip, stablesync} simply concatenate audio and visual features, which fails to effectively align the visual features with the audio in high-resolution conditions, leading to suboptimal synchronization. 
More recent work \cite{hallo} introduces a hierarchical cross-attention mechanism to augment the correlation between audio inputs and visual motions, but it remains unsuitable for real-time processing.

The specific operational steps of AVFM are straightforward. First, a $1\times1$ convolution is applied to the audio feature $f_a$ which is then treated as a filter to modulate the visual feature spectrum. Finally, the filtered feature $f_f$ is obtained by applying the inverse transformation to the filtered spectrum. In practice, the last modulated visual feature $f_{mod}^i$ is used as the visual input. Based on the previously defined terms and operations, the formulation steps for AVFM are as follows:
\begin{equation}
F_f = \mathcal{F}^{-1}(\mathcal{F}(f_{mod}) \odot \text{Conv}(f_a))
\end{equation}

The experiments demonstrate the effectiveness of AVFM, eliminating the need for a separate lip synchronization expert\cite{syncnet}. Such an expert often degrades visual quality and requires extensive datasets for pre-training.

\subsection{Loss Functions}
Given a synthesized talking head frame $I_o$ and a ground truth frame $I_{gt}$, three loss functions are used to train FREAK, including reconstruction loss, perceptual loss \cite{perceploss}, and frequency loss.

The L1 loss ensures pixel-level accuracy and structural consistency, as shown in \eqref{eq:l1_loss}. 
\begin{equation}
\mathcal{L}_{rec} =  \|I_o - I_{gt}\|_1,
\label{eq:l1_loss}
\end{equation}

The perceptual loss captures high-level semantic differences by comparing feature maps extracted from a pre-trained feature extractor, enhancing details and visual realism. It is defined as follows:
\begin{equation}
\mathcal{L}_{percep} = \| \mathcal{E}(I_o) - \mathcal{E}(I_{gt}) \|_2,
\label{eq:perceptual_loss}
\end{equation}
where $\mathcal{E}$ represents the feature extractor of VGG-19\cite{vgg19}.

Additionally, a frequency loss is applied, which is the L1 loss in the frequency domain between the spectrum of the generated image and the spectrum of the ground truth:
\begin{equation}
\mathcal{L}_{freq} =  \|\mathcal{F}(I_o) - \mathcal{F}(I_{gt})\|_1,
\label{eq:freq_loss}
\end{equation}

The final loss function for the model can be expressed as follows:
\begin{equation}
\mathcal{L} = \lambda_{rec} \mathcal{L}_{rec} + \lambda_{percep} \mathcal{L}_{percep} + \lambda_{freq} \mathcal{L}_{freq},
\label{eq:total_loss}
\end{equation}
where \(\lambda_{rec}\), \(\lambda_{percep}\), and \(\lambda_{freq}\) are weights 
of each loss.

\section{Experiments}
In this section, we conduct experiments to evaluate our model against state-of-the-art models. We first introduce various experimental settings, including datasets, comparison methods, evaluation metrics, and implementation details. Next, we evaluate our model from both quantitative and qualitative perspectives. Then, ablation experiments demonstrate the effectiveness of the modules, and finally, a user study presents a subjective comparison of the strengths and weaknesses among the models.
\subsection{Experimental Settings}
\subsubsection{Dataset}
We use data from \cite{adnerf} and \cite{hdtf}, selecting 7 videos containing 7 different IDs. The duration of each video ranges from 4 to 6 minutes. The first 80\% of each video is used as the training set, and the remaining 20\% is reserved for testing.

\subsubsection{Comparison Methods}
We compare our method with several state-of-the-art approaches including: Wav2Lip\cite{wav2lip}, DINet\cite{dinet}, MuseTalk\cite{musetalk}, IP-LAP\cite{iplap}, RAD-NeRF\cite{radnerf} and SadTalker\cite{sadtalker}. Among these, Wav2Lip, MuseTalk and RAD-NeRF are real-time methods. SadTalker belongs to one-shot methods.

\subsubsection{Comparison Metrics}
The proposed method is evaluated under three scenarios: reconstruction, cross-audio, and one-shot. In the reconstruction scenario, 
since ground truth of the same identity is available, Frechet Inception Distance (FID) \cite{fid} and Frechet Video Distance (FVD) \cite{fvd} are used for visual quality. Landmark Distance (LMD) \cite{lmd} is employed to evaluate mouth movement synchronization. In the cross-audio scenario, audio clips from \cite{dfrf}, including French and English, are used to drive all the models. As no ground truth of the same identity exists in this case, Lip Sync Error Distance (LSE-D) and Lip Sync Error Confidence (LSE-C) \cite{wav2lip} are adopted to evaluate the synchronization between lip movements and audio without references. 
The runtime for inferring a single frame serves as the running speed metric.

\begin{figure*}
    \centering
    \includegraphics[width=1\linewidth]{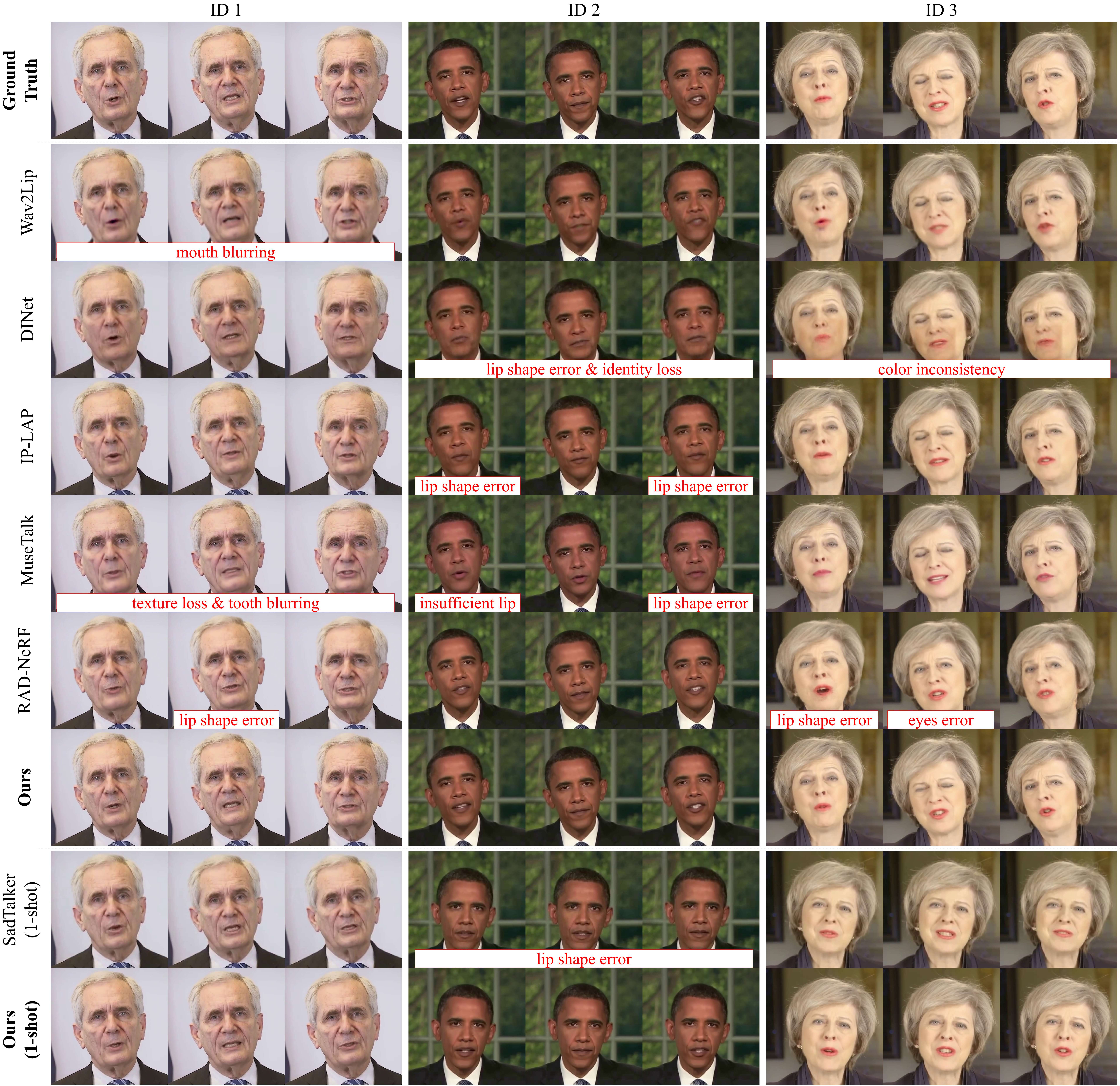}
    \caption{Qualitative Comparisons with State-of-the-Art Methods. Three identities, each with different speech content, are compared against Wav2Lip, DINet, IP-LAP, MuseTalk, RAD-NeRF and SadTalker. The results of SadTalker are used for comparison with those obtained by our method in the one-shot setting.}
    \label{fig:res}
\end{figure*}

\subsubsection{Implementation Details}
In our experiment, all videos are cropped to a resolution of $512\times512$ and set to 25 fps. During training, a random frame is selected from the video as the reference and the ground truth has a 50\% probability of masking either the lower half of the face or the entire head, which is used as the masked image.
We use hubert-large-ls960-ft for audio hubert feature extraction. For the extracted feature $f_{hu}\in\mathcal{R}^{t\times 2\times1024}$, 8 frames on both sides of the corresponding frame index are taken to form a combined feature of $f_{hu}\in\mathcal{R}^{16\times 2\times1024}$ (context length = 16), which is then reshaped into $\mathcal{R}^{32\times 32\times 32}$ as the input to the audio encoder. 
The network synthesizes videos frame by frame iteratively. In each iteration, the network takes one reference frame, one masked frame, and a segment of audio features mentioned above as input, and outputs one synchronized frame.

For the training setting, Adam with default parameters is used as the optimizer, and the learning rate is set to 0.001 with a batch size of 8. The loss function weights are set as $\lambda_{rec}=1$, $\lambda_{perc}=0.01$, and $\lambda_{freq}=1$, respectively. The entire model is trained from scratch.
All training and testing are performed on a single NVIDIA RTX A6000. 

\subsection{Quantitative Evaluation} 
As shown in Table \ref{quan}, our FREAK achieves the best performance in terms of FID, FVD and LMD. The performance on FVD indicates that our results are optimal in terms of both video quality and temporal consistency.
Since Wav2Lip is trained with a pre-trained SyncNet\cite{syncnet}, it attains exceptionally high scores on both LSE-D and LSE-C, even surpassing the ground truth. However, our method also outperforms most competing approaches in these metrics, demonstrating its strong synchronization capabilities. For LMD, which is calculated by subtracting the mean from each landmark coordinate to obtain a normalized distance, it reflects the accuracy of lip movements to some extent in the one-shot scenario. Compared to SadTalker, a single-image-driven method, our approach also demonstrates significant advantages in FID, FVD and LMD. Notably, SadTalker leverages a pre-trained Wav2Lip model for lip synthesis, resulting in reasonably high scores on the two synchronization metrics.

In terms of inference speed, we calculate the mean time consumption for each model to infer a single frame, excluding the time cost for pre-processing operations. Our method is faster than all other methods except Wav2Lip. Since Wav2Lip only produces very low-resolution outputs ($96\times96$) and does not employ intermediate representations or a multi-stage pipeline, its inference speed can achieve real-time performance. Our method adopts a similar network architecture to Wav2Lip but uses lightweight modules and operates in the Fourier space, so it is not surprising that it achieves real-time performance.

These findings underscore the robustness and versatility of our method under various experimental conditions.
\begin{figure}
    \centering
    \includegraphics[width=1\linewidth]{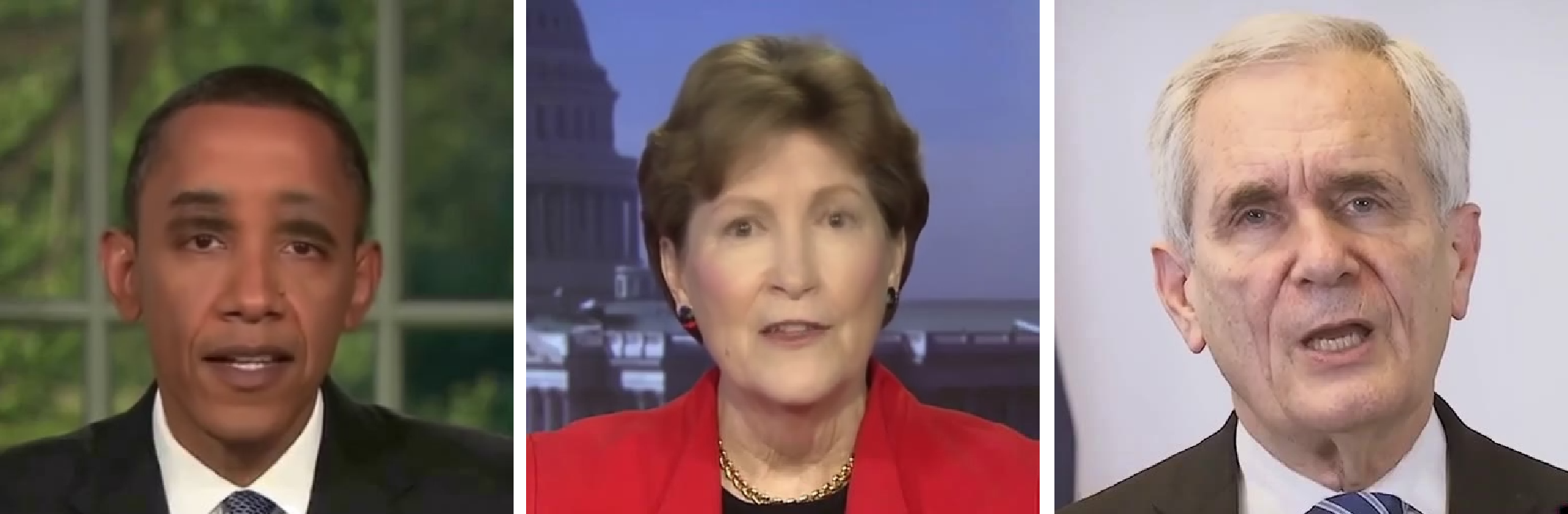}
    \caption{Results without both VEFM \& AVFM in the one-shot setting. The consistency of the head structure between frames is difficult to maintain, with noticeable jitter between frames. Some frames exhibit ghosting or distortion.}
    \label{fig:fail}
\end{figure}

\begin{figure}
    \centering
    \includegraphics[width=1\linewidth]{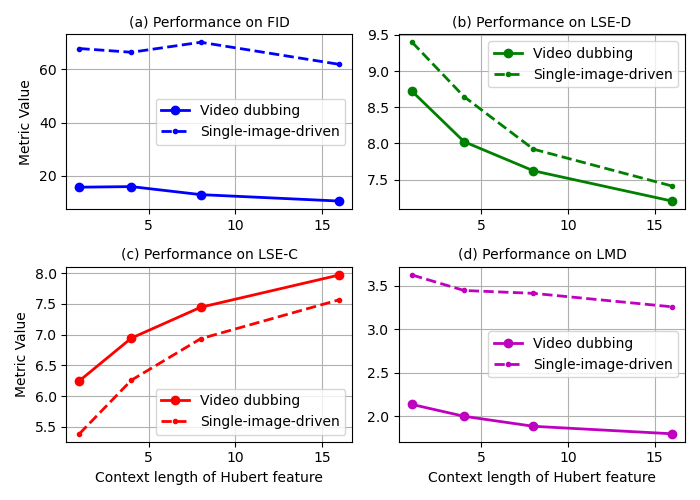}
    \caption{The impact of the context length of hubert features on the effect.}
    \label{fig:aud}
\end{figure}

\subsection{Qualitative Evaluation}
Our method demonstrates impressive performance in both visual quality and lip shape alignment, as illustrated in Fig. \ref{fig:res}. It can be observed that our approach achieves the best performance in terms of image resolution and texture preservation when compared to other methods, while also generating well-aligned lip shapes. 

In comparison to Wav2Lip, our results exhibit enhanced detail in the lip and teeth regions. For DINet, its results suffer from identity shift and color inconsistency issues. We can observe from the figure the loss of identity information for ID 2 and the lips not being open. IP-LAP can achieve fairly good results, but there are still some lip shape errors. Moreover, since it uses a two-stage pipeline and intermediate representations, it is not advantageous in terms of inference time. The problem with MuseTalk is the loss of skin texture, and there is a slight shift in identity information, along with some lip shape errors. RAD-NeRF can achieve good results in terms of image quality, identity preservation, and inference speed, but it is inferior to other methods in terms of lip-synchronization. Compared with SadTalker, their backgrounds are prone to movement and the lip-synchronization effect is not as good as that of our method.

\begin{table}[htbp]
\caption{Ablation for VEFM, AVFM and Frequency Loss}
\begin{center}
\begin{tabular}{|c|c|c|c|c|c|}
\hline
\textbf{Method} & \textbf{FID↓} & \textbf{LSE-D↓}& \textbf{LSE-C↑} & \textbf{LMD↓}\\
\hline
w/o VEFM & 12.294&7.275&7.779&1.889
\\
w/o AVFM & 12.547&7.571&7.509&1.896
\\
w/o VEFM\&AVFM & 17.344&8.177&6.665&2.032
\\
w/o $L_{freq}$ & 12.878&7.274&7.828&1.821
\\
FREAK & \textbf{10.514}&\textbf{7.202}&\textbf{7.972}&\textbf{1.800}
\\
\hline
w/o VEFM(1-shot) & 63.679&7.580&7.266&3.277
\\
w/o AVFM(1-shot) & 65.791&7.768&7.074&3.272
\\
w/o VEFM\&AVFM(1-shot) & 63.242&8.351&6.252&3.228
\\
w/o $L_{freq}$(1-shot) &66.845&7.496&7.388&3.304
\\
FREAK(1-shot) & \textbf{61.860}&\textbf{7.410}&\textbf{7.569}&\textbf{3.258}

\\
\hline
\end{tabular}
\label{ablation}
\end{center}
\end{table}

\begin{table*}[htbp]
\caption{User study.}
\begin{center}
\begin{tabular}{|c|c|c|c|c|c|c|c|}
\hline
\textbf{MOS on} & \textbf{Wav2Lip} & \textbf{DINet}& \textbf{IP-LAP} & \textbf{RAD-NeRF}& \textbf{MuseTalk}& \textbf{SadTalker} & \textbf{FREAK(Ours)}\\
\hline
Audio-Visual Sync. & 2.6& 3.4 & 2.867 & 2.6 & 3.133 &2 & \textbf{4.333}
\\
Video Quality & 1.333& 2.667 & 3.067 & 3.4 & 3.933 &2.8 & \textbf{4.733}
\\
Realness & 1.467 & 2.667 & 2.8 & 1.8 & 3.267 & 1.533 & \textbf{4.067}
\\
\hline
\end{tabular}
\label{user}
\end{center}
\end{table*}

\subsection{Ablation Study}


We conducted ablation experiments on several key components of our model, as shown in Table \ref{ablation}. For the variant without VEFM, the features encoded from the reference image and the mask image are directly concatenated, followed by a $1\times1$ convolution for channel transformation. Similarly, in the absence of AVFM, visual and audio features are directly concatenated. Moreover, we examined the impact of removing the frequency loss, and the effect of changing the context length of the input Hubert features. 

Both VEFM and AVFM contribute significantly to improving the model's performance. Removing AVFM leads to a substantial degradation in lip-sync accuracy, while removing VEFM also negatively affects overall performance, though its impact on lip-sync is less pronounced than that of AVFM. 

The removal of both modules has a substantial effect on the model, particularly in the one-shot setting. It was observed that without frequency-domain modulation, the one-shot setup results in head distortion and unstable jitter as shown in Fig. \ref{fig:fail}.
We speculate that this is because, in the full-head mask scenario, during the synthesis of part of the human head with dynamic audio features as conditions, the synthesized head is likely to deviate from the position of the head in the reference image. However, the spatial structural information learned by the network during head synthesis mainly comes from the reference image. Therefore, the synthesis results are not likely to deviate much from the head in the reference image. Such contradictions lead to the occurrence of head distortion or tearing and blurring.
Coupling features from the frequency domain enables the network to adaptively learn how to manipulate different spectra, and the two visual features can be coupled together harmoniously and naturally, thus obtaining natural head portraits.
This highlights the importance and necessity of frequency-domain modulation in the one-shot setting. Meanwhile, the use of frequency loss contributes to the model’s overall performance improvement.

Moreover, the context length of Hubert feature has a significant impact on synchronization and temporal consistency. When the context is insufficient, severe lip jitter occurs. This phenomenon diminishes as the context length increases. As shown in Fig. \ref{fig:aud}, the synchronization metrics improve significantly with the extension of the context length.

\subsection{User Study}
To provide a more comprehensive evaluation of the proposed model, we conduct user studies with 10 participants unrelated to this study on 21 videos generated by ours and the six other methods.
We adopt the widely used Mean Opinion Scores (MOS) rating protocol. Each participant is asked to rate from 1-5 for the talking portrait results based on three major aspects: audio-visual synchronization, video quality and realness. Audio-visual sync. focuses on the synchronization between audio and lip movements. Video quality is related to video clarity, facial clarity, and the preservation of textures. Realness is a more comprehensive aspect, indicating the authenticity of the video and determining whether it is AI-generated. The more natural the portrait's speech performance, the higher the score.
We collect the rating results and compute the average score of each method. The statistics are shown in Table \ref{user}.  FREAK surpasses previous methods in all evaluations. 

\section{Discussion and Conclusion}
We introduce frequency-domain learning to audio-driven talking head generation for the first time and propose FREAK, a novel method that generates high-fidelity and high-resolution talking portrait in real time  and supports seamless switching between one-shot mode and video dubbing. FREAK integrates and modulates features through two frequency-domain modulation modules, effectively improving the model performance and addressing the issues of head distortion in the single-image-driven scenario. Our experiments demonstrate the superiority of our method, offering valuable insights for future related work.

\textbf{Limitation.} If the training videos are too short or lack a sufficient variety of lip shapes, our method will not produce satisfactory results. Furthermore, since all training data uses front-facing talking portraits and experiments have not been conducted with side-facing profiles, this may lead to noticeable artifacts when the model encounters side-facing scenarios. 

\textbf{Broader Impacts.} Our method can generate natural and realistic talking head videos with few requirements. So it may bring concerns regarding potential misuse, especially in the creation of misleading
deepfake. Therefore, researchers must treat FREAK cautiously and rationally. We commit to sharing our results to further the development of deepfake detection tools. Meanwhile, the community should promote studies on face forge detection to automatically keep these videos away from social media.

\section{Acknowledgement}
This work was supported by the Fundamental Research Funds for the Central Universities of China.
\bibliographystyle{ACM-Reference-Format}
\bibliography{icmr2025references}


\begin{thebibliography}{55}


\ifx \showCODEN    \undefined \def \showCODEN     #1{\unskip}     \fi
\ifx \showISBNx    \undefined \def \showISBNx     #1{\unskip}     \fi
\ifx \showISBNxiii \undefined \def \showISBNxiii  #1{\unskip}     \fi
\ifx \showISSN     \undefined \def \showISSN      #1{\unskip}     \fi
\ifx \showLCCN     \undefined \def \showLCCN      #1{\unskip}     \fi
\ifx \shownote     \undefined \def \shownote      #1{#1}          \fi
\ifx \showarticletitle \undefined \def \showarticletitle #1{#1}   \fi
\ifx \showURL      \undefined \def \showURL       {\relax}        \fi
\providecommand\bibfield[2]{#2}
\providecommand\bibinfo[2]{#2}
\providecommand\natexlab[1]{#1}
\providecommand\showeprint[2][]{arXiv:#2}

\bibitem[Chen et~al\mbox{.}(2018)]%
        {lmd}
\bibfield{author}{\bibinfo{person}{Lele Chen}, \bibinfo{person}{Zhiheng Li}, \bibinfo{person}{Ross~K Maddox}, \bibinfo{person}{Zhiyao Duan}, {and} \bibinfo{person}{Chenliang Xu}.} \bibinfo{year}{2018}\natexlab{}.
\newblock \showarticletitle{Lip movements generation at a glance}. In \bibinfo{booktitle}{\emph{ECCV}}. \bibinfo{pages}{520--535}.
\newblock


\bibitem[Chen et~al\mbox{.}(2020)]%
        {ssdgan}
\bibfield{author}{\bibinfo{person}{Yuanqi Chen}, \bibinfo{person}{Ge Li}, \bibinfo{person}{Cece Jin}, \bibinfo{person}{Shan Liu}, {and} \bibinfo{person}{Thomas~H. Li}.} \bibinfo{year}{2020}\natexlab{}.
\newblock \showarticletitle{SSD-GAN: Measuring the Realness in the Spatial and Spectral Domains}. In \bibinfo{booktitle}{\emph{AAAI Conference on Artificial Intelligence}}.
\newblock
\urldef\tempurl%
\url{https://api.semanticscholar.org/CorpusID:228083505}
\showURL{%
\tempurl}


\bibitem[Chung and Zisserman(2017)]%
        {syncnet}
\bibfield{author}{\bibinfo{person}{Joon~Son Chung} {and} \bibinfo{person}{Andrew Zisserman}.} \bibinfo{year}{2017}\natexlab{}.
\newblock \showarticletitle{Out of time: automated lip sync in the wild}. In \bibinfo{booktitle}{\emph{Computer Vision--ACCV 2016 Workshops: ACCV 2016 International Workshops, Taipei, Taiwan, November 20-24, 2016, Revised Selected Papers, Part II 13}}. Springer, \bibinfo{pages}{251--263}.
\newblock


\bibitem[Daquan et~al\mbox{.}(2020)]%
        {mobilenet}
\bibfield{author}{\bibinfo{person}{Zhou Daquan}, \bibinfo{person}{Qibin Hou}, \bibinfo{person}{Yunpeng Chen}, \bibinfo{person}{Jiashi Feng}, {and} \bibinfo{person}{Shuicheng Yan}.} \bibinfo{year}{2020}\natexlab{}.
\newblock \showarticletitle{Rethinking Bottleneck Structure for Efficient Mobile Network Design}. In \bibinfo{booktitle}{\emph{European Conference on Computer Vision}}.
\newblock
\urldef\tempurl%
\url{https://api.semanticscholar.org/CorpusID:220363927}
\showURL{%
\tempurl}


\bibitem[Deng et~al\mbox{.}(2019)]%
        {3dmm}
\bibfield{author}{\bibinfo{person}{Yu Deng}, \bibinfo{person}{Jiaolong Yang}, \bibinfo{person}{Sicheng Xu}, \bibinfo{person}{Dong Chen}, \bibinfo{person}{Yunde Jia}, {and} \bibinfo{person}{Xin Tong}.} \bibinfo{year}{2019}\natexlab{}.
\newblock \showarticletitle{Accurate 3d face reconstruction with weakly-supervised learning: From single image to image set}. In \bibinfo{booktitle}{\emph{Proceedings of the IEEE/CVF conference on computer vision and pattern recognition workshops}}. \bibinfo{pages}{0--0}.
\newblock


\bibitem[Du et~al\mbox{.}(2023)]%
        {daetalker}
\bibfield{author}{\bibinfo{person}{Chenpeng Du}, \bibinfo{person}{Qi Chen}, \bibinfo{person}{Tianyu He}, \bibinfo{person}{Xu Tan}, \bibinfo{person}{Xie Chen}, \bibinfo{person}{Kai Yu}, \bibinfo{person}{Sheng Zhao}, {and} \bibinfo{person}{Jiang Bian}.} \bibinfo{year}{2023}\natexlab{}.
\newblock \showarticletitle{Dae-talker: High fidelity speech-driven talking face generation with diffusion autoencoder}. In \bibinfo{booktitle}{\emph{Proceedings of the 31st ACM MM}}. \bibinfo{pages}{4281--4289}.
\newblock


\bibitem[Fu et~al\mbox{.}(2025)]%
        {damc}
\bibfield{author}{\bibinfo{person}{Ao Fu}, \bibinfo{person}{Ziqi Ni}, {and} \bibinfo{person}{Yi Zhou}.} \bibinfo{year}{2025}\natexlab{}.
\newblock \showarticletitle{Dual Audio-Centric Modality Coupling for Talking Head Generation}.
\newblock \bibinfo{journal}{\emph{arXiv preprint arXiv:2503.22728}} (\bibinfo{year}{2025}).
\newblock


\bibitem[Gao et~al\mbox{.}({[n.\,d.]})]%
        {freqderain}
\bibfield{author}{\bibinfo{person}{Ning Gao}, \bibinfo{person}{Xingyu Jiang}, \bibinfo{person}{Xiuhui Zhang}, {and} \bibinfo{person}{Yue Deng}.} \bibinfo{year}{[n.\,d.]}\natexlab{}.
\newblock \showarticletitle{Efficient Frequency-Domain Image Deraining with Contrastive Regularization}.
\newblock  (\bibinfo{year}{[n.\,d.]}).
\newblock


\bibitem[Goyal et~al\mbox{.}(2023)]%
        {emogen}
\bibfield{author}{\bibinfo{person}{Sahil Goyal}, \bibinfo{person}{Sarthak Bhagat}, \bibinfo{person}{Shagun Uppal}, \bibinfo{person}{Hitkul Jangra}, \bibinfo{person}{Yi Yu}, \bibinfo{person}{Yifang Yin}, {and} \bibinfo{person}{Rajiv~Ratn Shah}.} \bibinfo{year}{2023}\natexlab{}.
\newblock \showarticletitle{Emotionally enhanced talking face generation}. In \bibinfo{booktitle}{\emph{Proceedings of the 1st International Workshop on Multimedia Content Generation and Evaluation: New Methods and Practice}}. \bibinfo{pages}{81--90}.
\newblock


\bibitem[Guibas et~al\mbox{.}(2021)]%
        {adaptivefourier}
\bibfield{author}{\bibinfo{person}{John Guibas}, \bibinfo{person}{Morteza Mardani}, \bibinfo{person}{Zongyi Li}, \bibinfo{person}{Andrew Tao}, \bibinfo{person}{Anima Anandkumar}, {and} \bibinfo{person}{Bryan Catanzaro}.} \bibinfo{year}{2021}\natexlab{}.
\newblock \showarticletitle{Adaptive fourier neural operators: Efficient token mixers for transformers}.
\newblock \bibinfo{journal}{\emph{arXiv preprint arXiv:2111.13587}} (\bibinfo{year}{2021}).
\newblock


\bibitem[Guo et~al\mbox{.}(2021)]%
        {adnerf}
\bibfield{author}{\bibinfo{person}{Yudong Guo}, \bibinfo{person}{Keyu Chen}, \bibinfo{person}{Sen Liang}, \bibinfo{person}{Yong-Jin Liu}, \bibinfo{person}{Hujun Bao}, {and} \bibinfo{person}{Juyong Zhang}.} \bibinfo{year}{2021}\natexlab{}.
\newblock \showarticletitle{Ad-nerf: Audio driven neural radiance fields for talking head synthesis}. In \bibinfo{booktitle}{\emph{ICCV}}. \bibinfo{pages}{5784--5794}.
\newblock


\bibitem[Heusel et~al\mbox{.}(2017)]%
        {fid}
\bibfield{author}{\bibinfo{person}{Martin Heusel}, \bibinfo{person}{Hubert Ramsauer}, \bibinfo{person}{Thomas Unterthiner}, \bibinfo{person}{Bernhard Nessler}, {and} \bibinfo{person}{Sepp Hochreiter}.} \bibinfo{year}{2017}\natexlab{}.
\newblock \showarticletitle{Gans trained by a two time-scale update rule converge to a local nash equilibrium}.
\newblock \bibinfo{journal}{\emph{NeurIPS}}  \bibinfo{volume}{30} (\bibinfo{year}{2017}).
\newblock


\bibitem[Hsu et~al\mbox{.}(2021)]%
        {hubert}
\bibfield{author}{\bibinfo{person}{Wei-Ning Hsu}, \bibinfo{person}{Benjamin Bolte}, \bibinfo{person}{Yao-Hung~Hubert Tsai}, \bibinfo{person}{Kushal Lakhotia}, \bibinfo{person}{Ruslan Salakhutdinov}, {and} \bibinfo{person}{Abdelrahman Mohamed}.} \bibinfo{year}{2021}\natexlab{}.
\newblock \showarticletitle{Hubert: Self-supervised speech representation learning by masked prediction of hidden units}.
\newblock \bibinfo{journal}{\emph{IEEE/ACM transactions on audio, speech, and language processing}}  \bibinfo{volume}{29} (\bibinfo{year}{2021}), \bibinfo{pages}{3451--3460}.
\newblock


\bibitem[Huang et~al\mbox{.}(2023)]%
        {aff}
\bibfield{author}{\bibinfo{person}{Zhipeng Huang}, \bibinfo{person}{Zhizheng Zhang}, \bibinfo{person}{Cuiling Lan}, \bibinfo{person}{Zheng-Jun Zha}, \bibinfo{person}{Yan Lu}, {and} \bibinfo{person}{Baining Guo}.} \bibinfo{year}{2023}\natexlab{}.
\newblock \showarticletitle{Adaptive frequency filters as efficient global token mixers}. In \bibinfo{booktitle}{\emph{ICCV}}. \bibinfo{pages}{6049--6059}.
\newblock


\bibitem[Jiang et~al\mbox{.}(2024)]%
        {loopy}
\bibfield{author}{\bibinfo{person}{Jianwen Jiang}, \bibinfo{person}{Chao Liang}, \bibinfo{person}{Jiaqi Yang}, \bibinfo{person}{Gaojie Lin}, \bibinfo{person}{Tianyun Zhong}, {and} \bibinfo{person}{Yanbo Zheng}.} \bibinfo{year}{2024}\natexlab{}.
\newblock \showarticletitle{Loopy: Taming audio-driven portrait avatar with long-term motion dependency}.
\newblock \bibinfo{journal}{\emph{arXiv preprint arXiv:2409.02634}} (\bibinfo{year}{2024}).
\newblock


\bibitem[Jiang et~al\mbox{.}(2021)]%
        {focalfreqloss}
\bibfield{author}{\bibinfo{person}{Liming Jiang}, \bibinfo{person}{Bo Dai}, \bibinfo{person}{Wayne Wu}, {and} \bibinfo{person}{Chen~Change Loy}.} \bibinfo{year}{2021}\natexlab{}.
\newblock \showarticletitle{Focal frequency loss for image reconstruction and synthesis}. In \bibinfo{booktitle}{\emph{ICCV}}. \bibinfo{pages}{13919--13929}.
\newblock


\bibitem[Johnson et~al\mbox{.}(2016)]%
        {perceploss}
\bibfield{author}{\bibinfo{person}{Justin Johnson}, \bibinfo{person}{Alexandre Alahi}, {and} \bibinfo{person}{Li Fei-Fei}.} \bibinfo{year}{2016}\natexlab{}.
\newblock \showarticletitle{Perceptual losses for real-time style transfer and super-resolution}. In \bibinfo{booktitle}{\emph{ECCV}}. Springer, \bibinfo{pages}{694--711}.
\newblock


\bibitem[Lanzino et~al\mbox{.}(2024)]%
        {realtimedeepfake}
\bibfield{author}{\bibinfo{person}{Romeo Lanzino}, \bibinfo{person}{Federico Fontana}, \bibinfo{person}{Anxhelo Diko}, \bibinfo{person}{Marco~Raoul Marini}, {and} \bibinfo{person}{Luigi Cinque}.} \bibinfo{year}{2024}\natexlab{}.
\newblock \showarticletitle{Faster Than Lies: Real-time Deepfake Detection using Binary Neural Networks}.
\newblock \bibinfo{journal}{\emph{2024 IEEE/CVF Conference on Computer Vision and Pattern Recognition Workshops (CVPRW)}} (\bibinfo{year}{2024}), \bibinfo{pages}{3771--3780}.
\newblock
\urldef\tempurl%
\url{https://api.semanticscholar.org/CorpusID:270357961}
\showURL{%
\tempurl}


\bibitem[Li et~al\mbox{.}(2023)]%
        {ernerf}
\bibfield{author}{\bibinfo{person}{Jiahe Li}, \bibinfo{person}{Jiawei Zhang}, \bibinfo{person}{Xiao Bai}, \bibinfo{person}{Jun Zhou}, {and} \bibinfo{person}{Lin Gu}.} \bibinfo{year}{2023}\natexlab{}.
\newblock \showarticletitle{Efficient region-aware neural radiance fields for high-fidelity talking portrait synthesis}. In \bibinfo{booktitle}{\emph{ICCV}}. \bibinfo{pages}{7568--7578}.
\newblock


\bibitem[Lin et~al\mbox{.}(2023)]%
        {freqdg}
\bibfield{author}{\bibinfo{person}{Shiqi Lin}, \bibinfo{person}{Zhizheng Zhang}, \bibinfo{person}{Zhipeng Huang}, \bibinfo{person}{Yan Lu}, \bibinfo{person}{Cuiling Lan}, \bibinfo{person}{Peng Chu}, \bibinfo{person}{Quanzeng You}, \bibinfo{person}{Jiang Wang}, \bibinfo{person}{Zicheng Liu}, \bibinfo{person}{Amey Parulkar}, {et~al\mbox{.}}} \bibinfo{year}{2023}\natexlab{}.
\newblock \showarticletitle{Deep frequency filtering for domain generalization}. In \bibinfo{booktitle}{\emph{CVPR}}. \bibinfo{pages}{11797--11807}.
\newblock


\bibitem[Lu et~al\mbox{.}(2021)]%
        {lsp}
\bibfield{author}{\bibinfo{person}{Yuanxun Lu}, \bibinfo{person}{Jinxiang Chai}, {and} \bibinfo{person}{Xun Cao}.} \bibinfo{year}{2021}\natexlab{}.
\newblock \showarticletitle{Live speech portraits: real-time photorealistic talking-head animation}.
\newblock \bibinfo{journal}{\emph{ACM Transactions on Graphics (TOG)}} \bibinfo{volume}{40}, \bibinfo{number}{6} (\bibinfo{year}{2021}), \bibinfo{pages}{1--17}.
\newblock


\bibitem[Ma et~al\mbox{.}(2023)]%
        {styletalk}
\bibfield{author}{\bibinfo{person}{Yifeng Ma}, \bibinfo{person}{Suzhen Wang}, \bibinfo{person}{Zhipeng Hu}, \bibinfo{person}{Changjie Fan}, \bibinfo{person}{Tangjie Lv}, \bibinfo{person}{Yu Ding}, \bibinfo{person}{Zhidong Deng}, {and} \bibinfo{person}{Xin Yu}.} \bibinfo{year}{2023}\natexlab{}.
\newblock \showarticletitle{Styletalk: One-shot talking head generation with controllable speaking styles}.
\newblock \bibinfo{journal}{\emph{arXiv preprint arXiv:2301.01081}} (\bibinfo{year}{2023}).
\newblock


\bibitem[Masi et~al\mbox{.}(2020)]%
        {freqtwobranch}
\bibfield{author}{\bibinfo{person}{Iacopo Masi}, \bibinfo{person}{Aditya Killekar}, \bibinfo{person}{Royston~Marian Mascarenhas}, \bibinfo{person}{Shenoy~Pratik Gurudatt}, {and} \bibinfo{person}{Wael AbdAlmageed}.} \bibinfo{year}{2020}\natexlab{}.
\newblock \showarticletitle{Two-branch Recurrent Network for Isolating Deepfakes in Videos}.
\newblock \bibinfo{journal}{\emph{ArXiv}}  \bibinfo{volume}{abs/2008.03412} (\bibinfo{year}{2020}).
\newblock
\urldef\tempurl%
\url{https://api.semanticscholar.org/CorpusID:221090663}
\showURL{%
\tempurl}


\bibitem[Mildenhall et~al\mbox{.}(2020)]%
        {nerf}
\bibfield{author}{\bibinfo{person}{Ben Mildenhall}, \bibinfo{person}{Pratul~P. Srinivasan}, \bibinfo{person}{Matthew Tancik}, \bibinfo{person}{Jonathan~T. Barron}, \bibinfo{person}{Ravi Ramamoorthi}, {and} \bibinfo{person}{Ren Ng}.} \bibinfo{year}{2020}\natexlab{}.
\newblock \showarticletitle{NeRF}.
\newblock \bibinfo{journal}{\emph{Commun. ACM}}  \bibinfo{volume}{65} (\bibinfo{year}{2020}), \bibinfo{pages}{99 -- 106}.
\newblock
\urldef\tempurl%
\url{https://api.semanticscholar.org/CorpusID:213175590}
\showURL{%
\tempurl}


\bibitem[Mukhopadhyay et~al\mbox{.}(2024)]%
        {diff2lip}
\bibfield{author}{\bibinfo{person}{Soumik Mukhopadhyay}, \bibinfo{person}{Saksham Suri}, \bibinfo{person}{Ravi~Teja Gadde}, {and} \bibinfo{person}{Abhinav Shrivastava}.} \bibinfo{year}{2024}\natexlab{}.
\newblock \showarticletitle{Diff2lip: Audio conditioned diffusion models for lip-synchronization}. In \bibinfo{booktitle}{\emph{Proceedings of the IEEE/CVF Winter Conference on Applications of Computer Vision}}. \bibinfo{pages}{5292--5302}.
\newblock


\bibitem[Pitas(2000)]%
        {dip}
\bibfield{author}{\bibinfo{person}{I Pitas}.} \bibinfo{year}{2000}\natexlab{}.
\newblock \showarticletitle{Digital Image Processing Algorithms and Applications}.
\newblock \bibinfo{journal}{\emph{John Wiley \& Sons Inc google schola}}  \bibinfo{volume}{2} (\bibinfo{year}{2000}), \bibinfo{pages}{133--138}.
\newblock


\bibitem[Prajwal et~al\mbox{.}(2020)]%
        {wav2lip}
\bibfield{author}{\bibinfo{person}{KR Prajwal}, \bibinfo{person}{Rudrabha Mukhopadhyay}, \bibinfo{person}{Vinay~P Namboodiri}, {and} \bibinfo{person}{CV Jawahar}.} \bibinfo{year}{2020}\natexlab{}.
\newblock \showarticletitle{A lip sync expert is all you need for speech to lip generation in the wild}. In \bibinfo{booktitle}{\emph{Proceedings of the 28th ACM MM}}. \bibinfo{pages}{484--492}.
\newblock


\bibitem[Rahaman et~al\mbox{.}(2018)]%
        {spectralbias}
\bibfield{author}{\bibinfo{person}{Nasim Rahaman}, \bibinfo{person}{Aristide Baratin}, \bibinfo{person}{Devansh Arpit}, \bibinfo{person}{Felix Dr{\"a}xler}, \bibinfo{person}{Min Lin}, \bibinfo{person}{Fred~A. Hamprecht}, \bibinfo{person}{Yoshua Bengio}, {and} \bibinfo{person}{Aaron~C. Courville}.} \bibinfo{year}{2018}\natexlab{}.
\newblock \showarticletitle{On the Spectral Bias of Neural Networks}. In \bibinfo{booktitle}{\emph{International Conference on Machine Learning}}.
\newblock
\urldef\tempurl%
\url{https://api.semanticscholar.org/CorpusID:53012119}
\showURL{%
\tempurl}


\bibitem[Rao et~al\mbox{.}(2021)]%
        {globalfilter}
\bibfield{author}{\bibinfo{person}{Yongming Rao}, \bibinfo{person}{Wenliang Zhao}, \bibinfo{person}{Zheng Zhu}, \bibinfo{person}{Jiwen Lu}, {and} \bibinfo{person}{Jie Zhou}.} \bibinfo{year}{2021}\natexlab{}.
\newblock \showarticletitle{Global filter networks for image classification}.
\newblock \bibinfo{journal}{\emph{NeurIPS}}  \bibinfo{volume}{34} (\bibinfo{year}{2021}), \bibinfo{pages}{980--993}.
\newblock


\bibitem[Shen et~al\mbox{.}(2022)]%
        {dfrf}
\bibfield{author}{\bibinfo{person}{Shuai Shen}, \bibinfo{person}{Wanhua Li}, \bibinfo{person}{Zheng Zhu}, \bibinfo{person}{Yueqi Duan}, \bibinfo{person}{Jie Zhou}, {and} \bibinfo{person}{Jiwen Lu}.} \bibinfo{year}{2022}\natexlab{}.
\newblock \showarticletitle{Learning dynamic facial radiance fields for few-shot talking head synthesis}. In \bibinfo{booktitle}{\emph{ECCV}}. Springer, \bibinfo{pages}{666--682}.
\newblock


\bibitem[Shen et~al\mbox{.}(2023)]%
        {difftalk}
\bibfield{author}{\bibinfo{person}{Shuai Shen}, \bibinfo{person}{Wenliang Zhao}, \bibinfo{person}{Zibin Meng}, \bibinfo{person}{Wanhua Li}, \bibinfo{person}{Zheng Zhu}, \bibinfo{person}{Jie Zhou}, {and} \bibinfo{person}{Jiwen Lu}.} \bibinfo{year}{2023}\natexlab{}.
\newblock \showarticletitle{Difftalk: Crafting diffusion models for generalized talking head synthesis}.
\newblock \bibinfo{journal}{\emph{arXiv preprint arXiv:2301.03786}} (\bibinfo{year}{2023}).
\newblock


\bibitem[Simonyan and Zisserman(2014)]%
        {vgg19}
\bibfield{author}{\bibinfo{person}{Karen Simonyan} {and} \bibinfo{person}{Andrew Zisserman}.} \bibinfo{year}{2014}\natexlab{}.
\newblock \showarticletitle{Very Deep Convolutional Networks for Large-Scale Image Recognition}.
\newblock \bibinfo{journal}{\emph{CoRR}}  \bibinfo{volume}{abs/1409.1556} (\bibinfo{year}{2014}).
\newblock
\urldef\tempurl%
\url{https://api.semanticscholar.org/CorpusID:14124313}
\showURL{%
\tempurl}


\bibitem[Song et~al\mbox{.}(2020)]%
        {everybodytalk}
\bibfield{author}{\bibinfo{person}{Linsen Song}, \bibinfo{person}{Wayne Wu}, \bibinfo{person}{Chen Qian}, \bibinfo{person}{Ran He}, {and} \bibinfo{person}{Chen~Change Loy}.} \bibinfo{year}{2020}\natexlab{}.
\newblock \showarticletitle{Everybody’s Talkin’: Let Me Talk as You Want}.
\newblock \bibinfo{journal}{\emph{IEEE Transactions on Information Forensics and Security}}  \bibinfo{volume}{17} (\bibinfo{year}{2020}), \bibinfo{pages}{585--598}.
\newblock
\urldef\tempurl%
\url{https://api.semanticscholar.org/CorpusID:210701290}
\showURL{%
\tempurl}


\bibitem[Stypu{\l}kowski et~al\mbox{.}(2024)]%
        {diffusedhead}
\bibfield{author}{\bibinfo{person}{Micha{\l} Stypu{\l}kowski}, \bibinfo{person}{Konstantinos Vougioukas}, \bibinfo{person}{Sen He}, \bibinfo{person}{Maciej Zi{\k{e}}ba}, \bibinfo{person}{Stavros Petridis}, {and} \bibinfo{person}{Maja Pantic}.} \bibinfo{year}{2024}\natexlab{}.
\newblock \showarticletitle{Diffused heads: Diffusion models beat gans on talking-face generation}. In \bibinfo{booktitle}{\emph{Proceedings of the IEEE/CVF Winter Conference on Applications of Computer Vision}}. \bibinfo{pages}{5091--5100}.
\newblock


\bibitem[Sun et~al\mbox{.}(2022)]%
        {avcat}
\bibfield{author}{\bibinfo{person}{Yasheng Sun}, \bibinfo{person}{Hang Zhou}, \bibinfo{person}{Kaisiyuan Wang}, \bibinfo{person}{Qianyi Wu}, \bibinfo{person}{Zhibin Hong}, \bibinfo{person}{Jingtuo Liu}, \bibinfo{person}{Errui Ding}, \bibinfo{person}{Jingdong Wang}, \bibinfo{person}{Ziwei Liu}, {and} \bibinfo{person}{Koike Hideki}.} \bibinfo{year}{2022}\natexlab{}.
\newblock \showarticletitle{Masked Lip-Sync Prediction by Audio-Visual Contextual Exploitation in Transformers}.
\newblock \bibinfo{journal}{\emph{SIGGRAPH Asia 2022 Conference Papers}} (\bibinfo{year}{2022}).
\newblock
\urldef\tempurl%
\url{https://api.semanticscholar.org/CorpusID:254070649}
\showURL{%
\tempurl}


\bibitem[Tan et~al\mbox{.}(2024)]%
        {freqawaredetection}
\bibfield{author}{\bibinfo{person}{Chuangchuang Tan}, \bibinfo{person}{Yao Zhao}, \bibinfo{person}{Shikui Wei}, \bibinfo{person}{Guanghua Gu}, \bibinfo{person}{Ping Liu}, {and} \bibinfo{person}{Yunchao Wei}.} \bibinfo{year}{2024}\natexlab{}.
\newblock \showarticletitle{Frequency-Aware Deepfake Detection: Improving Generalizability through Frequency Space Learning}.
\newblock \bibinfo{journal}{\emph{ArXiv}}  \bibinfo{volume}{abs/2403.07240} (\bibinfo{year}{2024}).
\newblock
\urldef\tempurl%
\url{https://api.semanticscholar.org/CorpusID:268890333}
\showURL{%
\tempurl}


\bibitem[Tang et~al\mbox{.}(2022)]%
        {radnerf}
\bibfield{author}{\bibinfo{person}{Jiaxiang Tang}, \bibinfo{person}{Kaisiyuan Wang}, \bibinfo{person}{Hang Zhou}, \bibinfo{person}{Xiaokang Chen}, \bibinfo{person}{Dongliang He}, \bibinfo{person}{Tianshu Hu}, \bibinfo{person}{Jingtuo Liu}, \bibinfo{person}{Gang Zeng}, {and} \bibinfo{person}{Jingdong Wang}.} \bibinfo{year}{2022}\natexlab{}.
\newblock \showarticletitle{Real-time neural radiance talking portrait synthesis via audio-spatial decomposition}.
\newblock \bibinfo{journal}{\emph{arXiv preprint arXiv:2211.12368}} (\bibinfo{year}{2022}).
\newblock


\bibitem[Tatsunami and Taki(2024)]%
        {dynamicfilter}
\bibfield{author}{\bibinfo{person}{Yuki Tatsunami} {and} \bibinfo{person}{Masato Taki}.} \bibinfo{year}{2024}\natexlab{}.
\newblock \showarticletitle{Fft-based dynamic token mixer for vision}. In \bibinfo{booktitle}{\emph{AAAI}}, Vol.~\bibinfo{volume}{38}. \bibinfo{pages}{15328--15336}.
\newblock


\bibitem[Tian et~al\mbox{.}(2025)]%
        {emo}
\bibfield{author}{\bibinfo{person}{Linrui Tian}, \bibinfo{person}{Qi Wang}, \bibinfo{person}{Bang Zhang}, {and} \bibinfo{person}{Liefeng Bo}.} \bibinfo{year}{2025}\natexlab{}.
\newblock \showarticletitle{EMO: Emote Portrait Alive Generating Expressive Portrait Videos with Audio2Video Diffusion Model Under Weak Conditions}. In \bibinfo{booktitle}{\emph{ECCV}}. Springer, \bibinfo{pages}{244--260}.
\newblock


\bibitem[Unterthiner et~al\mbox{.}(2019)]%
        {fvd}
\bibfield{author}{\bibinfo{person}{Thomas Unterthiner}, \bibinfo{person}{Sjoerd van Steenkiste}, \bibinfo{person}{Karol Kurach}, \bibinfo{person}{Rapha{\"e}l Marinier}, \bibinfo{person}{Marcin Michalski}, {and} \bibinfo{person}{Sylvain Gelly}.} \bibinfo{year}{2019}\natexlab{}.
\newblock \showarticletitle{FVD: A new Metric for Video Generation}. In \bibinfo{booktitle}{\emph{DGS@ICLR}}.
\newblock
\urldef\tempurl%
\url{https://api.semanticscholar.org/CorpusID:198489709}
\showURL{%
\tempurl}


\bibitem[Wang et~al\mbox{.}(2024)]%
        {freqmotionmag}
\bibfield{author}{\bibinfo{person}{Fei Wang}, \bibinfo{person}{Dan Guo}, \bibinfo{person}{Kun Li}, \bibinfo{person}{Zhun Zhong}, {and} \bibinfo{person}{Meng Wang}.} \bibinfo{year}{2024}\natexlab{}.
\newblock \showarticletitle{Frequency decoupling for motion magnification via multi-level isomorphic architecture}. In \bibinfo{booktitle}{\emph{CVPR}}. \bibinfo{pages}{18984--18994}.
\newblock


\bibitem[Wang et~al\mbox{.}(2023)]%
        {talklip}
\bibfield{author}{\bibinfo{person}{Jiadong Wang}, \bibinfo{person}{Xinyuan Qian}, \bibinfo{person}{Malu Zhang}, \bibinfo{person}{Robby~T Tan}, {and} \bibinfo{person}{Haizhou Li}.} \bibinfo{year}{2023}\natexlab{}.
\newblock \showarticletitle{Seeing What You Said: Talking Face Generation Guided by a Lip Reading Expert}. In \bibinfo{booktitle}{\emph{CVPR}}. \bibinfo{pages}{14653--14662}.
\newblock


\bibitem[Wang et~al\mbox{.}(2022)]%
        {freqdeepfake}
\bibfield{author}{\bibinfo{person}{Junke Wang}, \bibinfo{person}{Zuxuan Wu}, \bibinfo{person}{Wenhao Ouyang}, \bibinfo{person}{Xintong Han}, \bibinfo{person}{Jingjing Chen}, \bibinfo{person}{Yu-Gang Jiang}, {and} \bibinfo{person}{Ser-Nam Li}.} \bibinfo{year}{2022}\natexlab{}.
\newblock \showarticletitle{M2tr: Multi-modal multi-scale transformers for deepfake detection}. In \bibinfo{booktitle}{\emph{Proceedings of the 2022 international conference on multimedia retrieval}}. \bibinfo{pages}{615--623}.
\newblock


\bibitem[Wei et~al\mbox{.}(2024)]%
        {aniportrait}
\bibfield{author}{\bibinfo{person}{Huawei Wei}, \bibinfo{person}{Zejun Yang}, {and} \bibinfo{person}{Zhisheng Wang}.} \bibinfo{year}{2024}\natexlab{}.
\newblock \showarticletitle{Aniportrait: Audio-driven synthesis of photorealistic portrait animation}.
\newblock \bibinfo{journal}{\emph{arXiv preprint arXiv:2403.17694}} (\bibinfo{year}{2024}).
\newblock


\bibitem[Xiong et~al\mbox{.}(2024)]%
        {segtalker}
\bibfield{author}{\bibinfo{person}{Lingyu Xiong}, \bibinfo{person}{Xize Cheng}, \bibinfo{person}{Jintao Tan}, \bibinfo{person}{Xianjia Wu}, \bibinfo{person}{Xiandong Li}, \bibinfo{person}{Lei Zhu}, \bibinfo{person}{Fei Ma}, \bibinfo{person}{Minglei Li}, \bibinfo{person}{Huang Xu}, {and} \bibinfo{person}{Zhihui Hu}.} \bibinfo{year}{2024}\natexlab{}.
\newblock \showarticletitle{SegTalker: Segmentation-based Talking Face Generation with Mask-guided Local Editing}. In \bibinfo{booktitle}{\emph{Proceedings of the 32nd ACM MM}}. \bibinfo{pages}{3170--3179}.
\newblock


\bibitem[Xu et~al\mbox{.}(2024)]%
        {hallo}
\bibfield{author}{\bibinfo{person}{Mingwang Xu}, \bibinfo{person}{Hui Li}, \bibinfo{person}{Qingkun Su}, \bibinfo{person}{Hanlin Shang}, \bibinfo{person}{Liwei Zhang}, \bibinfo{person}{Ce Liu}, \bibinfo{person}{Jingdong Wang}, \bibinfo{person}{Yao Yao}, {and} \bibinfo{person}{Siyu Zhu}.} \bibinfo{year}{2024}\natexlab{}.
\newblock \showarticletitle{Hallo: Hierarchical audio-driven visual synthesis for portrait image animation}.
\newblock \bibinfo{journal}{\emph{arXiv preprint arXiv:2406.08801}} (\bibinfo{year}{2024}).
\newblock


\bibitem[Yaman et~al\mbox{.}({[n.\,d.]})]%
        {stablesync}
\bibfield{author}{\bibinfo{person}{Dogucan Yaman}, \bibinfo{person}{Fevziye~Irem Eyiokur}, \bibinfo{person}{Leonard B{\"a}rmann}, \bibinfo{person}{Haz{\i}m~Kemal Ekenel}, {and} \bibinfo{person}{Alexander Waibel}.} \bibinfo{year}{[n.\,d.]}\natexlab{}.
\newblock \showarticletitle{Audio-driven Talking Face Generation with Stabilized Synchronization Loss}.
\newblock  (\bibinfo{year}{[n.\,d.]}).
\newblock


\bibitem[Yang et~al\mbox{.}(2022)]%
        {freqmotion}
\bibfield{author}{\bibinfo{person}{Guang Yang}, \bibinfo{person}{Wu Liu}, \bibinfo{person}{Xinchen Liu}, \bibinfo{person}{Xiaoyan Gu}, \bibinfo{person}{Juan Cao}, {and} \bibinfo{person}{Jintao Li}.} \bibinfo{year}{2022}\natexlab{}.
\newblock \showarticletitle{Delving into the Frequency: Temporally Consistent Human Motion Transfer in the Fourier Space}.
\newblock \bibinfo{journal}{\emph{Proceedings of the 30th ACM International Conference on Multimedia}} (\bibinfo{year}{2022}).
\newblock
\urldef\tempurl%
\url{https://api.semanticscholar.org/CorpusID:251979541}
\showURL{%
\tempurl}


\bibitem[Yu et~al\mbox{.}(2019)]%
        {gatedconv}
\bibfield{author}{\bibinfo{person}{Jiahui Yu}, \bibinfo{person}{Zhe Lin}, \bibinfo{person}{Jimei Yang}, \bibinfo{person}{Xiaohui Shen}, \bibinfo{person}{Xin Lu}, {and} \bibinfo{person}{Thomas~S Huang}.} \bibinfo{year}{2019}\natexlab{}.
\newblock \showarticletitle{Free-form image inpainting with gated convolution}. In \bibinfo{booktitle}{\emph{ICCV}}. \bibinfo{pages}{4471--4480}.
\newblock


\bibitem[Zhang et~al\mbox{.}(2023a)]%
        {sadtalker}
\bibfield{author}{\bibinfo{person}{Wenxuan Zhang}, \bibinfo{person}{Xiaodong Cun}, \bibinfo{person}{Xuan Wang}, \bibinfo{person}{Yong Zhang}, \bibinfo{person}{Xi Shen}, \bibinfo{person}{Yu Guo}, \bibinfo{person}{Ying Shan}, {and} \bibinfo{person}{Fei Wang}.} \bibinfo{year}{2023}\natexlab{a}.
\newblock \showarticletitle{SadTalker: Learning Realistic 3D Motion Coefficients for Stylized Audio-Driven Single Image Talking Face Animation}. In \bibinfo{booktitle}{\emph{CVPR}}. \bibinfo{pages}{8652--8661}.
\newblock


\bibitem[Zhang et~al\mbox{.}(2020)]%
        {hdtf}
\bibfield{author}{\bibinfo{person}{Xi Zhang}, \bibinfo{person}{Xiaolin Wu}, \bibinfo{person}{Xinliang Zhai}, \bibinfo{person}{Xianye Ben}, {and} \bibinfo{person}{Chengjie Tu}.} \bibinfo{year}{2020}\natexlab{}.
\newblock \showarticletitle{Davd-net: Deep audio-aided video decompression of talking heads}. In \bibinfo{booktitle}{\emph{CVPR}}. \bibinfo{pages}{12335--12344}.
\newblock


\bibitem[Zhang et~al\mbox{.}(2024)]%
        {musetalk}
\bibfield{author}{\bibinfo{person}{Yue Zhang}, \bibinfo{person}{Minhao Liu}, \bibinfo{person}{Zhaokang Chen}, \bibinfo{person}{Bin Wu}, \bibinfo{person}{Yubin Zeng}, \bibinfo{person}{Chao Zhan}, \bibinfo{person}{Yingjie He}, \bibinfo{person}{Junxin Huang}, {and} \bibinfo{person}{Wenjiang Zhou}.} \bibinfo{year}{2024}\natexlab{}.
\newblock \showarticletitle{MuseTalk: Real-Time High Quality Lip Synchronization with Latent Space Inpainting}.
\newblock \bibinfo{journal}{\emph{arXiv preprint arXiv:2410.10122}} (\bibinfo{year}{2024}).
\newblock


\bibitem[Zhang et~al\mbox{.}(2023b)]%
        {dinet}
\bibfield{author}{\bibinfo{person}{Zhimeng Zhang}, \bibinfo{person}{Zhipeng Hu}, \bibinfo{person}{Wenjin Deng}, \bibinfo{person}{Changjie Fan}, \bibinfo{person}{Tangjie Lv}, {and} \bibinfo{person}{Yu Ding}.} \bibinfo{year}{2023}\natexlab{b}.
\newblock \showarticletitle{Dinet: Deformation inpainting network for realistic face visually dubbing on high resolution video}. In \bibinfo{booktitle}{\emph{AAAI}}, Vol.~\bibinfo{volume}{37}. \bibinfo{pages}{3543--3551}.
\newblock


\bibitem[Zhong et~al\mbox{.}(2023)]%
        {iplap}
\bibfield{author}{\bibinfo{person}{Weizhi Zhong}, \bibinfo{person}{Chaowei Fang}, \bibinfo{person}{Yinqi Cai}, \bibinfo{person}{Pengxu Wei}, \bibinfo{person}{Gangming Zhao}, \bibinfo{person}{Liang Lin}, {and} \bibinfo{person}{Guanbin Li}.} \bibinfo{year}{2023}\natexlab{}.
\newblock \showarticletitle{Identity-Preserving Talking Face Generation with Landmark and Appearance Priors}. In \bibinfo{booktitle}{\emph{CVPR}}. \bibinfo{pages}{9729--9738}.
\newblock


\bibitem[Zhou et~al\mbox{.}(2020)]%
        {makeittalk}
\bibfield{author}{\bibinfo{person}{Yang Zhou}, \bibinfo{person}{Xintong Han}, \bibinfo{person}{Eli Shechtman}, \bibinfo{person}{Jose Echevarria}, \bibinfo{person}{Evangelos Kalogerakis}, {and} \bibinfo{person}{Dingzeyu Li}.} \bibinfo{year}{2020}\natexlab{}.
\newblock \showarticletitle{Makelttalk: speaker-aware talking-head animation}.
\newblock \bibinfo{journal}{\emph{ACM Transactions On Graphics (TOG)}} \bibinfo{volume}{39}, \bibinfo{number}{6} (\bibinfo{year}{2020}), \bibinfo{pages}{1--15}.
\newblock


\end{thebibliography}

\end{document}